\documentclass[letterpaper, 10 pt, conference]{ieeeconf}  
\IEEEoverridecommandlockouts                              

\usepackage{graphics} 
\usepackage{epsfig} 
\usepackage{mathptmx} 
\usepackage{times} 
\usepackage{amsmath} 
\usepackage{amssymb}  

\usepackage{graphicx}
\usepackage{caption}
\usepackage{subcaption}
\usepackage{multirow}
\usepackage{adjustbox}
\usepackage{colortbl}
\usepackage{hhline}
\usepackage{relsize}
\usepackage{verbatim}
\usepackage{tabularx}
\usepackage{booktabs}
\usepackage{hyperref}
\hypersetup{colorlinks, citecolor=green,hypertexnames=true}

\begin{document}

\title{\LARGE \bf
Learning Panoptic Segmentation from Instance Contours
}
\author{Sumanth Chennupati$^{1}$, Venkatraman Narayanan$^2$, Ganesh Sistu$^3$, Senthil Yogamani$^3$ and Samir A Rawashdeh$^1$
\thanks{$^{1}$Department of Electrical and Computer Engineering, University of Michigan-Dearborn, 4901 Evergreen Rd, Dearborn, MI USA. {\tt\small \{schenn,srawa\}@umich.edu}}%
\thanks{$^{2}$University of Maryland-College Park, College Park, MD USA. {\tt\small vnarayan@terpmail.umd.edu}}%
\thanks{$^{3}$Valeo Vision Systems, Tuam, Ireland
{\tt\small \{ganesh.sistu, senthil.yogamani\}@valeo.com}}%
}


\maketitle

\begin{abstract}
Panoptic Segmentation aims to provide an understanding of background (stuff) and instances of objects (things) at a pixel level. It combines the separate tasks of semantic segmentation (pixel level classification) and instance segmentation to build a single unified scene understanding task. Typically, panoptic segmentation is derived by combining semantic and instance segmentation tasks that are learned separately or jointly (multi-task networks). In general, instance segmentation networks are built by adding a foreground mask estimation layer on top of object detectors or using instance clustering methods that assign a pixel to an instance center. In this work, we present a fully convolution neural network that learns instance segmentation from semantic segmentation and instance contours (boundaries of things). Instance contours along with semantic segmentation yield a boundary aware semantic segmentation of things. Connected component labeling on these results produces instance segmentation. We merge semantic and instance segmentation results to output panoptic segmentation. We evaluate our proposed method on the CityScapes dataset to demonstrate qualitative and quantitative performances along with several ablation studies. Our overview video can be accessed from \url{https://youtu.be/wBtcxRhG3e0}.
\end{abstract}


\section{Introduction}

Panoptic segmentation \cite{kirillov2019panopticfpn, kirillov2019panoptic} offers ultimate understanding of a scene by providing joint semantic and instance level predictions of background and objects at a pixel level. Panoptic segmentation is usually achieved by combining outputs from semantic segmentation  and instance segmentation. Examples where panoptic segmentation offers unprecedented advantage over standalone semantic or instance segmentation solutions include collective knowledge of distinct objects and drivable area around a self-driving car \cite{petrovai2019multi, de2019single}, semantic and instance level details of cancerous cells in digital pathology \cite{zhang2018panoptic}, understanding of background and different individuals in a frame to enhance smartphone photography. Multi-task learning networks  that jointly perform semantic \cite{ravikumar2021omnidet, sistu2019neurall, Chennupati_2019_CVPR_Workshops, auxnet} and instance segmentation  \cite{kirillov2019panopticfpn, petrovai2019multi} accelerated progress of panoptic segmentation in terms of accuracy and computational efficiency compared to traditional methods that use naive fusion of predictions from independent semantic and instance segmentation networks to derive panoptic segmentation output \cite{kirillov2019panoptic}.

\begin{figure}[tb]
\centering
\vspace{0.25cm}
\includegraphics[trim=15 0 15 0, clip, width=\columnwidth]{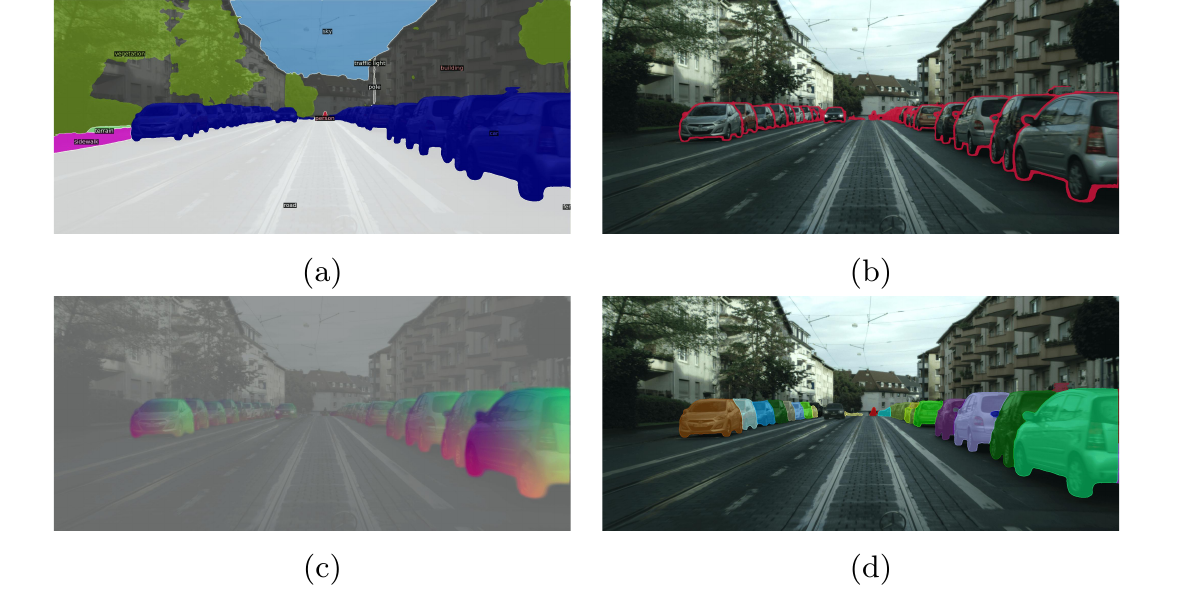}


\caption{(a) Semantic segmentation, (b) Instance contour segmentation, (c) Instance center regression and (d) Instance segmentation} 
\vspace{-0.5cm}
\label{fig:intro}
\end{figure}

Instance segmentation is typically achieved in two major ways,  
1) Foreground mask estimation of objects detected by an object detection model \cite{kirillov2019panopticfpn, li2018weakly, he2017mask} or 2) Clustering based instance assignment methods \cite{kendall2018multi, cheng2019panoptic}. Recently, single stage instance segmentation methods have been developed \cite{Xie_2020_CVPR, hurtik2020poly}. These major approaches use fully convolution networks so that they can be trained in an end-to-end fashion. 

Semantic segmentation is a mature task which is well explored in literature relatively to panoptic segmentation. We make an observation that panoptic segmentation can be obtained from semantic segmentation by additionally estimating instance separating contours. 
Naively, the instance separating contours can be an additional class in the segmentation task. In practice, it  is difficult to get good performance for this class. 
It is illustrated in Figure \ref{fig:intro} where segmentation (a) and instance contour segmentation (b) contains all the information to obtain panoptic segmentation. The minimal contours needed are contours which separate two instances of the same object. However, these contours do not have sufficient information to be learnt on its own and thus we use the entire instance contours.  

\begin{figure}[tb]
\centering
\vspace{0.25cm}
\includegraphics[width=\columnwidth]{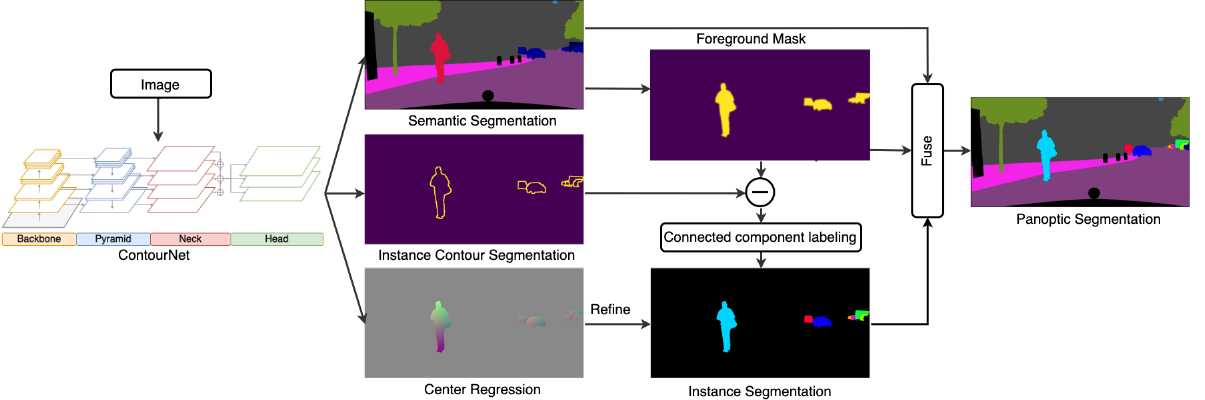}
\caption{We present a network that learns panoptic segmentation from semantic segmentation and instance contours (boundaries of things). We use a shared convolution neural network to predict semantic segmentation, instance contours and center regression. Instance contours along with semantic segmentation yield a boundary aware semantic segmentation of things. Connected component labeling on these results produce instance segmentation and eventually panoptic segmentation.}
\vspace{-0.5cm}
\label{fig:teaser}
\end{figure}

In this work, we present a multi-task learning network as shown in Figure \ref{fig:teaser} that learns semantic segmentation, instance contours and center regression  results. Our instance contours along with semantic segmentation guide us to derive instance segmentation and eventually panoptic segmentation. We also estimate a confidence score for each instance. Our instance contour segmentation network is a binary segmentation network that predicts instance boundaries between objects that belong to a same category. 
Compared to semantic edge detection networks \cite{yu2017casenet, Acuna_2019_CVPR} our instance contour estimation does not ignore boundaries between instances of a same category. 
We split large instances or merge small instances, using 2-d offsets to an instance center predicted by instance center regression at a pixel level.

We hope that our idea encourages a new direction in the research of panoptic segmentation which ultimately leads to learning of instance separating contours within the segmentation task.
The main contributions of this paper include:
\begin{enumerate}
    \item A novel method to learn panoptic segmentation and instance segmentation from semantic segmentation and instance contours.
    \item An instance contour segmentation network that learns boundaries between objects of same semantic category.
\end{enumerate}

\section{Related Work}
\label{sec:relatedwork}
Scene understanding \cite{hoiem2015guest} has witnessed tremendous progress over the past decade with introduction of Convolution Neural Networks \cite{lecun1999object, krizhevsky2012imagenet, he2016deep} that aided in development of semantic segmentation (pixel wise classification) and instance segmentation (pixel level recognition of distinct objects). Panoptic segmentation \cite{kirillov2019panoptic}, a joint semantic and instance segmentation has provided complete scene understanding by categorizing a pixel into distinct categories and instances. On the other hand, semantic edge detection \cite{yu2017casenet} has been widely used to learn boundaries between semantic classes.

\subsection{Semantic Segmentation}
\label{sec:background-semseg}
Few years ago semantic segmentation \cite{siam2017deep} was considered a challenging problem. With the help of Fully convolutional neural networks (FCNs) \cite{noh2015learning} development of accurate and efficient solutions were made possible \cite{siam2018comparative, siam2018rtseg}. 

Several enhancements were made to push the performance of semantic segmentation higher by making improvements to encoder and decoder in FCNs. Dilated residual convolutions \cite{chen2017deeplab}, Feature pyramid networks \cite{kirillov2019panopticfpn, lin2017feature}, Spatial pyramid pooling \cite{he2015spatial} etc. are examples of improvements made to encoder while U-Net \cite{unet}, Densely connected CRFs \cite{chen2017deeplab, zheng2015conditional} are examples of improvements made to decoder. We use a combination of feature pyramid networks and a light weight asymmetric decoder presented by Kirillov et al. \cite{kirillov2019panopticfpn} to learn semantic segmentation. 

\subsection{Instance Segmentation}
\label{sec:background-instseg}
In instance segmentation, an object instance(id) is assigned to every pixel for every known object within an image.
Two-stage methods like Mask R-CNN \cite{he2017mask} involve proposal generation from object detection 
followed by mask generation using a foreground/background binary segmentation network. These methods dominate the state of the art in instance segmentation but incur a relatively higher computational cost. 
Using YOLO \cite{redmon2016you}, SSD \cite{liu2016ssd} and other light wight object detector compared to Faster R-CNN \cite{ren2015faster} may seem promising but they still posses inevitable additional compute in generating object proposals followed by mask generation. 

Other approaches in instance segmentation range from clustering of instance embedding \cite{liang2017proposal} to prediction of instance centers using offset regression \cite{kendall2018multi, cheng2019panoptic}. 
These methods appear logically straight forward but are lagging behind in terms of accuracy and computational efficiency. The major drawback with these methods is usage of compute intensive clustering methods like OPTICS \cite{ankerst1999optics}, DBSCAN \cite{ester1996density} etc. In contrast to these methods, we derive instance segmentation from semantic segmentation using instance contours (boundaries of things).

\subsection{Semantic Edge Detection}

Semantic edge detection (SED) \cite{yu2017casenet, yang2016object} differs from edge detection \cite{van1989nonlinear} by predicting edges that belong to semantic class boundaries. In SED, edges/boundaries that separate segments of one category from another are predicted whereas, in edge detection every edge is detected based on image gradients. 
Holistically-nested edge detection (HED) \cite{xie2015holistically} is one of the first CNN based edge detection method.
Later, several methods were proposed to address different challenges in edge detection that include prediction of crisp boundaries \cite{Acuna_2019_CVPR, Deng_2018_ECCV}, selection of intermediate feature maps and choices of supervision on these feature maps \cite{liu2017richer, bertasius2015deepedge}. It is important to note that these methods ignore the boundaries between instances of objects that belong to same semantic category. 

Deep contour \cite{deepcontour} is similar to our method where instance contours are used generate instance segmentation. Deep Snake \cite{Peng_2020_CVPR} recently proposed to predict instance contours by learning contours from object detection. They replace foreground mask estimation for objects with contours to derive instance segmentation. Our instance contour segmentation however is a single stage method that directly estimates contours using a binary segmentation network.

\subsection{Panoptic Segmentation }
\label{sec:background-panseg}

Panoptic segmentation \cite{kirillov2019panoptic}  
combines semantic segmentation and instance segmentation together to provide class category and instance id for every pixel within an image. Recent works \cite{kirillov2019panopticfpn, cheng2019panoptic, petrovai2019multi} use a shared backbone and predict panoptic segmentation by fusing output from semantic and instance segmentation branches. Almost every work so far uses an FCN based semantic segmentation branch with variations including usage of dilated convolutions \cite{cheng2019panoptic} or feature pyramid networks \cite{kirillov2019panopticfpn}. However, choices of instance segmentation branch can vary as discussed in Section \ref{sec:background-instseg}.

Major challenge in generating panoptic segmentation output is merging conflicting outputs from semantic segmentation and instance branches. For example, semantic segmentation can predict that a pixel might belong to car class while instance segmentation branch may predict the same pixel as person class. 
Several methods \cite{petrovai2019multi} were proposed to handle the conflicts in a better and learned fashion. 
Our methods propose to derive instance segmentation from semantic segmentation using instance contours. Therefore, our method does not require a conflict resolution policy like other existing methods.




\section{Proposed Method}
\label{sec:methods}

\begin{figure}[ht]
\centering
\includegraphics[width=\columnwidth]{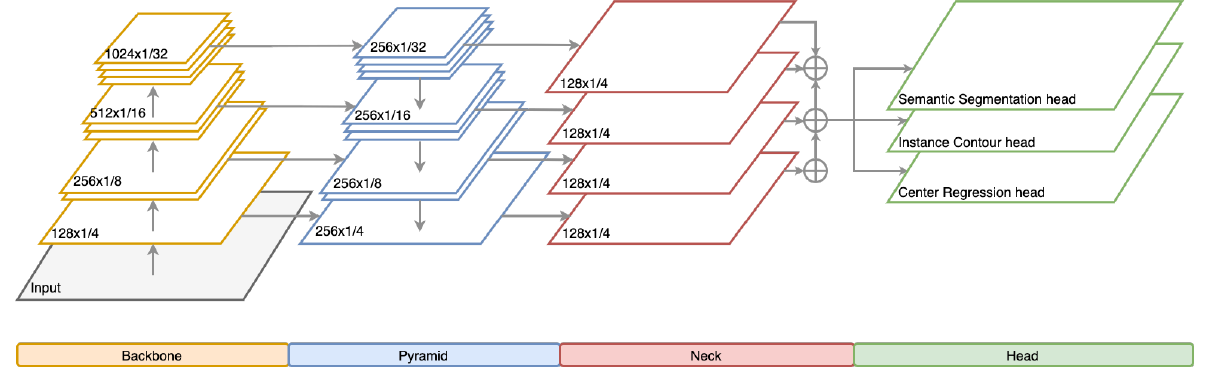}
\caption{ Proposed model architecture with CNN backbone. Multi-scale features from the backbone are fed to a feature pyramid network and then to upsampling neck followed by a prediction head. Our network has three heads for semantic segmentation, instance contour segmentation and center regression tasks. Separate necks can be used for different heads/tasks as needed} 
\label{fig:arch}
\vspace{-15pt}
\end{figure}

Our proposed method is a multi-task neural network with several shared convolution layers and multiple output heads that predict semantic segmentation, instance contours and center regression. As shown in Figure \ref{fig:arch}, a common ResNet \cite{he2016deep} backbone outputs multi-scale feature maps that are processed by a top-down feature pyramid network \cite{lin2017feature}. These feature maps from different levels are upsampled to a common scale through a series of 1x1 convolutions and combined before making output predictions. We refer the upsampling stages as necks and prediction layers as heads. 

Outputs from instance contour and semantic segmentation branches are combined to generate instance segmentation. We refine instance segmentation output using center regression results. Later, we simply merge semantic and instance segmentation outputs to generate panoptic segmentation. 

\subsection{Model Architecture}
\label{sec:methods-arch}

We begin with introducing our shared backbone that outputs multi-scale feature maps as shown in Figure \ref{fig:arch}. Our backbone uses ResNet \cite{he2016deep} as the encoder that outputs multiple scales of feature maps \{1/4, 1/8, 1/16, 1/32\} w.r.t to input image. Our pyramid is built using Feature pyramid network (FPN) \cite{lin2017feature} which consumes feature maps (scales 1/4 to 1/32) from backbone in a top-down fashion and outputs feature maps with 256 channels maintaining their input scale. Feature maps from pyramid are then passed through a series of 1x1 convolutions and are upsampled to 1/4 scale using 2-d bi-linear interpolation in the neck layers as proposed in \cite{kirillov2019panopticfpn}. These layers have 128 dims at each level. We add these feature maps from different levels and pass to prediction heads. Our semantic segmentation head contains 1$\times$1 convolution layer with \textit{k} filters (to output \textit{k} output maps for \textit{k} classes) followed by a 4x upsampling. 
We perform softmax activation followed by an argmax function on the \textit{k} output maps to derive full resolution semantic segmentation output. Our instance contour estimation head is similar to semantic segmentation head except it has 1 output feature map and a sigmoid activation instead of softmax. Our center regression head has two output channels that predict offsets from instance center in x and y axis, and does not have any special activation function.


\subsection{Loss functions}
\label{sec:methods-loss}
We discuss the explicit loss functions defined for semantic segmentation and instance contour branches. We chose cross-entropy loss for semantic segmentation. In Equation \ref{eq:seg-loss}, $L_{semantic}$ is segmentation loss over \textit{k} classes for all pixels in the image, where $p_i$ is the prediction probability and $\hat{y}_i$ is ground truth that indicates whether pixel belongs to class $i$.
\begin{equation}
    \label{eq:seg-loss}
    \displaystyle L_{semantic} = -{\mathlarger\sum_{i}^k} \hat{y_i} \cdot log(p_i)
\end{equation}

For instance contours, we chose weighted binary cross entropy loss
\cite{yu2017casenet} as shown in Equation \ref{eq:multi-label-loss}, where $\beta$ is the ratio of non edge pixels to total pixels $n$ in the image. $p_i$ is the probability that current pixel is an edge and $\hat{y}_i$ is ground truth which indicates whether pixel i is an edge.
\begin{equation}
    \label{eq:multi-label-loss}
        L_{wBCE} = -{\mathlarger\sum_{i}^n} \Big \langle \beta \cdot \hat{y}_i \cdot log(p_i)  + (1-\beta)\cdot(1-\hat{y}_i)\cdot log(1- p_i) \Big \rangle
\end{equation}
We add Huber loss ($\delta$ = 0.3):
\begin{equation}
    \label{eq:huber-loss}
    \displaystyle L_{Huber} = 
 \begin{cases} 
 0.5 \cdot (p_i - \hat{y}_i)^2, &\quad |p_i - \hat{y}_i| \leq \delta \\ 
 \delta \cdot (p_i - \hat{y}_i) - 0.5 \cdot \delta^2,   &\quad otherwise 
 \end{cases} 
\end{equation}
and NMS Loss \{$L_{NMS} = - \sum_{c} log(h) $\} \cite{Acuna_2019_CVPR} terms to contour loss to predict thin and crisp boundaries. We compute softmax response $h$ along the normal direction of boundary pixels $c$ as described in \cite{Acuna_2019_CVPR}. 
For center regression, we use Huber loss to compute error between $y$, predicted offsets and $\hat{y}$, ground truth offsets with $\delta$ = 1. 

Our total loss function is a weight combination of semantic loss, contour losses and center regression loss.
\begin{equation}
    \label{eq:total-loss}
\displaystyle L_{total} = \lambda_1 \cdot L_{semantic} + \lambda_2 \cdot L_{contour} + \lambda_3 \cdot L_{center}    
\end{equation}
where $L_{contour}$ is defined as:
\begin{equation}
    \label{eq:contour-loss}
    \displaystyle L_{contour} = L_{wBCE} + L_{Huber} + L_{NMS}
\end{equation}
We chose $\lambda_1$, $\lambda_2$, and $\lambda_3$ as 1, 50 and 0.1 for our experiments.

\begin{figure}[tb]
\centering
\includegraphics[trim=10 15 10 0, clip, width=\columnwidth]{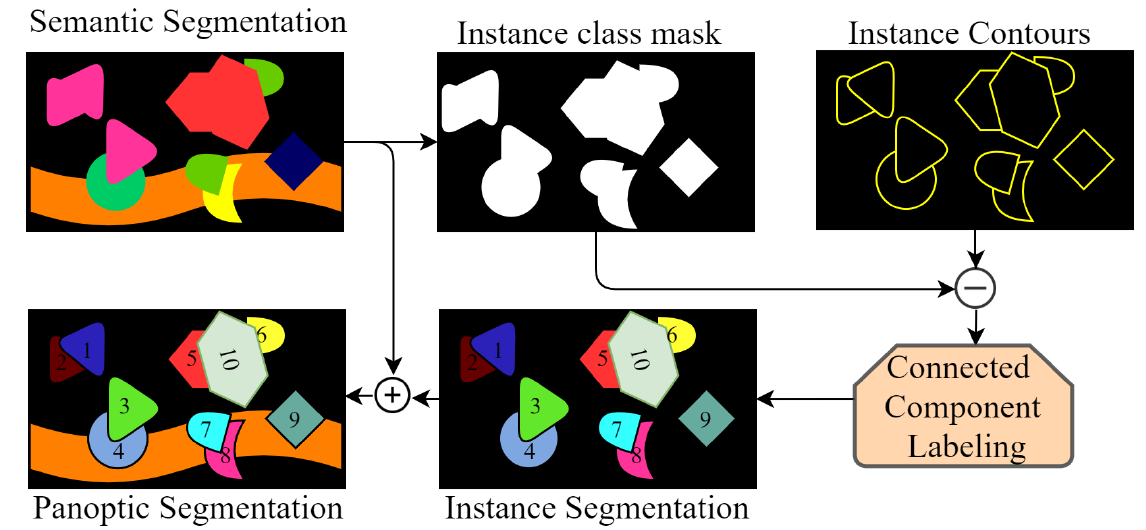}
\caption{Illustrative flow diagram of proposed algorithm that learns panoptic segmentation from semantic segmentation and instance contours} 
\vspace{-0.5cm}
\label{fig:arch-block}
\end{figure}



\subsection{Instance segmentation}

Our instance segmentation is derived from semantic segmentation unlike any other instance segmentation methods as shown in Figure \ref{fig:arch-block}. As a first step, we generate a binary mask by searching for instance classes in semantic segmentation which we refer to as instance class mask. We subtract instance contours (generated from instance contour segmentation head) from instance class mask to derive boundary aware instance class mask. Using connected component labeling \cite{samet1988efficient}, we derive unique instances from boundary aware instance class mask. We map the semantic segmentation output to the instance generated. We assign the most frequent label found inside an instance as its category and average the softmax predictions over the area of an instance to generate confidence for an instance.

\subsection{Refining Instance Segmentation}
\label{sec:methods-refine}

We refine instance segmentation output using center regression results. Our refinement consists of mainly 2 stages: Split and Merge. We estimate centroids predicted by center regression head. We cluster the centroid predictions using DBSCAN in an instance and split them if distinct centroids are found. If distance between two centroids is at least 20 pixels (eps), we declare them as distinct. For a 1024$\times$2048 image, we believed that 20 pixels is relatively enough to distinguish smaller instances. Our clustering stage does not require large computational complexity like other methods \cite{liang2017proposal, kendall2018multi, cheng2019panoptic} since we perform clustering within instances that are much smaller compared to performing clustering on entire image. 

After the instances are split, we estimate mean centroids for every instance using offsets predicted by center regression head. If the mean centroids are closer than 20 pixels in euclidean distance, we merge those instances. Later, we remove all instances that have an area lower than a minimum area threshold. We assign these pixels to instances whose centroids are closest to the centroids derived from offsets predicted by the center regression head. 

\subsection{Panoptic Segmentation}
Panoptic segmentation is now obtained by simply merging output from semantic segmentation and instance segmentation. As discussed in Section \ref{sec:background-panseg} we do not use a conflict resolution since our instance segmentation is a byproduct of our semantic segmentation. Thus, we will never have conflicting predictions.

\section{Experiments, Results and Discussion}
\label{sec:experiments}

\begin{figure*}[tb]
\centering

\includegraphics[trim=0 0 0 2180, clip, width=\textwidth]{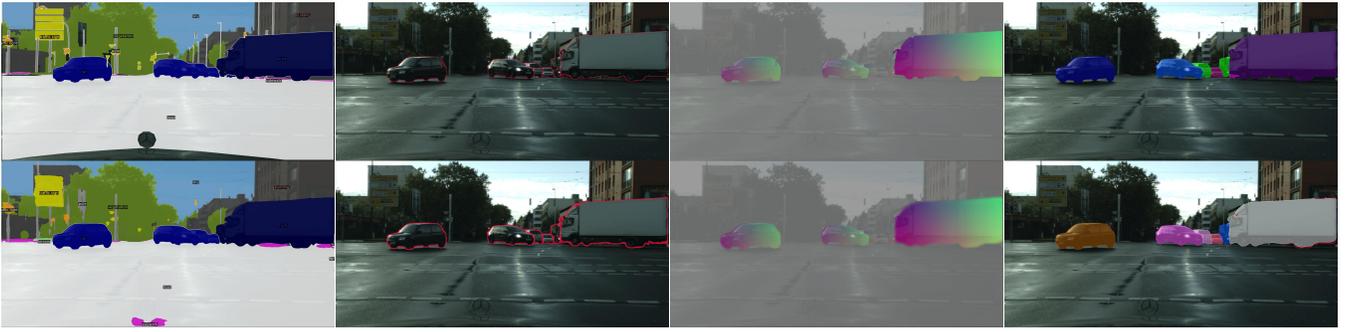}

\caption{Qualitative Results on Cityscapes val \cite{cordts2016cityscapes} dataset obtained with ResNet-50 \cite{he2016deep} encoder using separate necks, wBCE + Huber loss combination, split and merge refinement with min Instance area = 300 pixels. Instance contours ground truth are generated with dilation rate = 2. From Left to Right: Semantic Segmentation, Instance contour Segmentation, Center Regression, Instance Segmentation. Top: Ground Truth, Bottom: Prediction. Predicted Contours are thicker than ground truth.} 
\vspace{-0.5cm}
\label{fig:contour-results}
\end{figure*}

In this section, we demonstrate the performance of our proposed methods for panoptic segmentation on Cityscapes \cite{cordts2016cityscapes} dataset specifically on the validation split. We also present the performance of our semantic segmentation and instance segmentation results. 

\subsection{Experimental Setup}
Cityscapes \cite{cordts2016cityscapes} is an automotive scene understanding dataset with 2975/500 train/val images at 1024$\times$2048 resolution. This dataset contains labels for semantic, instance and panoptic segmentation tasks. We derive labels for our instance contour task by applying a contour detection algorithm on instance ground truth masks. We dilate the resulting contours to derive thick contours and serve them as ground truth for our instance contour segmentation task. Cityscapes dataset has 19 semantic object categories out of which 8 categories are provided with instance masks.

We train our network on full resolution images with a batch size of 4 images. We use Group Normalization \cite{wu2018group} which is effective for lower batch sizes. We use an SGD optimizer with learning rate = $0.005$, momentum = 0.9, weight decay = $10^{-4}$.
We initialize our ResNet encoders with pre-trained ImageNet \cite{deng2009imagenet} weights and train our networks for 48000 iterations.

We measure the performance of semantic segmentation using mean intersection over union (mIoU), instance segmentation using mean average precision (mAP) and panoptic segmentation using panoptic quality (PQ) \cite{kirillov2019panoptic}, segmentation quality (SQ) and recognition quality (RQ) metrics. 

\begin{table}[b]
\vspace{-0.5cm}

\centering
\resizebox{\columnwidth}{!}{
\begin{tabular}{@{}ccc|cc|ccc@{}}
\toprule
\multicolumn{3}{c|}{Contour Loss} & \multicolumn{5}{c}{Performance} \\
wBCE & Huber & NMS & AP & PQ & PQ\textsuperscript{Th} & SQ\textsuperscript{Th} & RQ\textsuperscript{Th} \\ \midrule
  \checkmark  &       &   & 16.0  & 43.9    & 25.0   &  72.6  &  33.3  \\ 
  \checkmark  &    \checkmark     &    &   \textbf{24.3}   &  \textbf{ 47.8} &\textbf{33.2}   & \textbf{76.3}   & \textbf{42.9}  \\  
  \checkmark  &       &    \checkmark  &  18.9 &  44.6  & 26.1   & 74.3  & 35.3\\ 
  \checkmark  &    \checkmark    &   \checkmark   & 23.3     &  46.7  &  32.4  &  76.1  & 42.1\\ \bottomrule
\end{tabular}
}
\caption{Instance and Panoptic Segmentation results on Cityscapes val \cite{cordts2016cityscapes} dataset for different loss functions used to represent instance contour loss. wBCE = weighted binary cross entropy, AP = average precision, PQ = panoptic quality. PQ\textsuperscript{Th}, SQ\textsuperscript{Th}, and RQ\textsuperscript{Th} represent panoptic, segmentation and  recognition qualities of instance objects.}

\label{tab:lossfn-ablation}
\end{table}

\subsection{Ablation experiments}
\label{sec:results}

\subsubsection{Instance contour segmentation loss function}
As mentioned before, we aim to predict thin and crisp instance contours. We study different loss functions discussed in Section \ref{sec:methods-loss} by evaluating the performance of instance and panoptic segmentation as shown in Table \ref{tab:lossfn-ablation}. We used ResNet-50 encoder as our backbone and separate heads with a common neck as discussed in Section \ref{sec:methods-arch}. 

\begin{figure}[tb]
\vspace{0.25cm}
\centering

\includegraphics[trim=150 70 10 0, clip, width=\columnwidth]{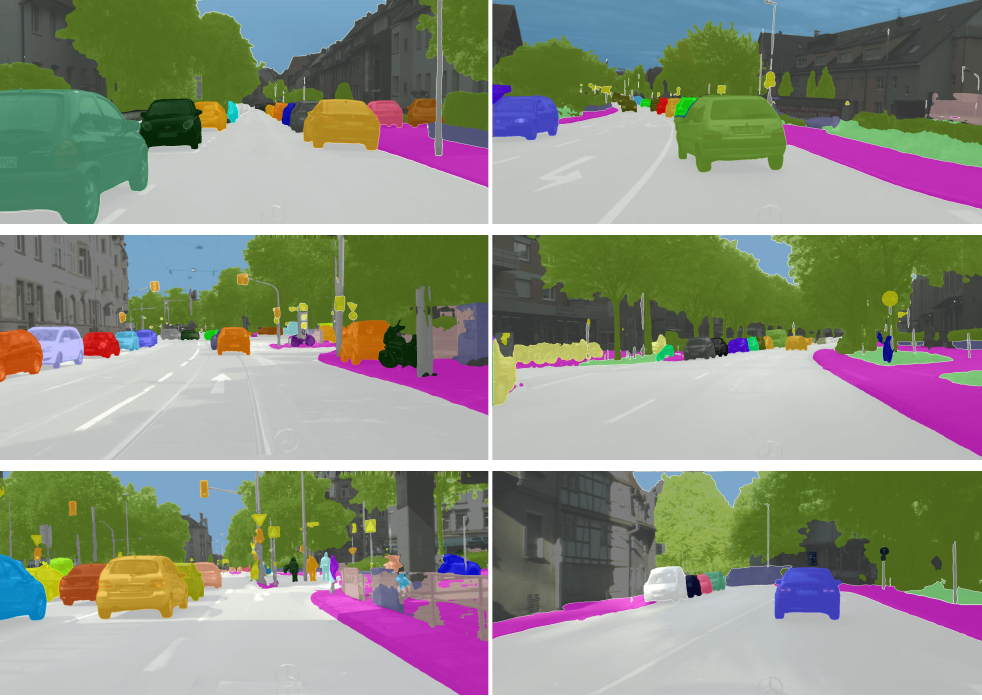}




\caption{Panoptic Segmentation Results on Cityscapes val \cite{cordts2016cityscapes} dataset. Results obtained with ResNet-50 \cite{he2016deep} encoder using a separate neck architecture, wBCE + Huber loss combination, split and merge refinement with min Instance area = 300 pixels. Instance contours ground truth are generated with dilation rate = 2.} 
\vspace{-0.7cm}
\label{fig:pan-results}
\end{figure}

We observed that the use of Huber and NMS loss function have improved the performance of instance and panoptic segmentation results. The weighted binary cross entropy combined with the Huber loss is the best combination we found. We use this combination for the rest of the experiments in the paper. Qualitative results in Figure \ref{fig:contour-results} demonstrate that the contours generated are thin and crisp when the above combination is used. In Figure \ref{fig:pan-results}, we demonstrate more qualitative results of the panoptic segmentation.

\begin{table}[htb]
\vspace{-0.25cm}
\centering
\begin{tabular}{@{}ccc|cc|ccc@{}}
\toprule
\multicolumn{3}{c|}{Dilation Rate}   & AP & PQ  & PQ\textsuperscript{Th} & SQ\textsuperscript{Th} & RQ\textsuperscript{Th} \\ \midrule
\multicolumn{3}{c|}{1}               &  24.1 & 46.0  & 30.5   &  73.1  &  40.6  \\
\multicolumn{3}{c|}{2}  &   \textbf{24.3}   &   \textbf{47.8} &\textbf{33.2}   & \textbf{76.3}  & \textbf{42.9}  \\
\multicolumn{3}{c|}{3} & 22.6    &   46.6 & 32.0   & 75.6   &  41.7  \\ \bottomrule
\end{tabular}
\caption{Performance of instance and panoptic segmentation on Cityscapes val \cite{cordts2016cityscapes} dataset when different dilation rates were used to generate ground truth instance contours. Increasing the dilation rate, increases the thickness of the ground truth instance contours.}
\label{tab:dilation-rate-ablation}
\vspace{-0.5cm}
\end{table}

\subsubsection{Instance contour ground truth dilation rate}
We generate our ground truth instance contours by applying a contour detection algorithm on instance masks provided for different objects in cityscapes dataset. Number of edge pixels are comparatively lower than non edge pixels in our contour segmentation problem. We can alleviate this class imbalance using appropriate loss functions as discussed in Section \ref{sec:methods-loss} or by dilating the the contours and increasing their thickness. In Table \ref{tab:dilation-rate-ablation}, we evaluate the performance of instance and panoptic segmentation for different dilation rates.

We observed that when appropriate loss combination is used, the dilation rate does not have a significant impact on the performance. However, increasing the dilation rate from 2 to 3 decreases the performance. We use a dilation rate of 2 to generate our ground truth contours for all other experiments. 

\begin{table}[htb]
\vspace{-0.15cm}
\centering
\begin{tabular}{cc|cc|ccc}
\toprule
\multicolumn{2}{c|}{Refine} & \multicolumn{5}{c}{Performance} \\
Split & Merge  & AP & PQ & PQ\textsuperscript{Th} & SQ\textsuperscript{Th} & RQ\textsuperscript{Th} \\ \midrule
      &      &  24.0   & 47.1 & 33.0 & 75.6   & 42.7   \\ 
  \checkmark  & &  24.2 &  47.7  &  33.1 & 76.1 & \textbf{42.9}\\
  \checkmark   &   \checkmark      & \textbf{24.3}  &  \textbf{47.8}  & \textbf{33.2}  & \textbf{76.3} & \textbf{42.9}\\
  \bottomrule
\end{tabular}
\caption{Evaluation of instance and panoptic segmentation on Cityscapes val \cite{cordts2016cityscapes} dataset before and after refinement using offsets predicted by center regression results.}
\label{tab:refine-ablation}
\vspace{-0.25cm}
\end{table}

\subsubsection{Refining Instance Segmentation}
As discussed in Section \ref{sec:methods-refine}, we refine our instance segmentation output using center regression results. We evaluate the effects of split and merge components in our refinement process in Table \ref{tab:refine-ablation} and evaluate the effect of min instance area in Table \ref{tab:min-pix-area-ablation}.

\begin{table}[htb]
\centering
\begin{tabular}{@{}ccc|cc|ccc@{}}
\toprule
\multicolumn{3}{c|}{min Instance Area}   & AP  & PQ &  PQ\textsuperscript{Th} & SQ\textsuperscript{Th} & RQ\textsuperscript{Th} \\ \midrule
\multicolumn{3}{c|}{1}  &  10.0    &  40.6    &  17.6   & 75.5  & 23.1    \\
\multicolumn{3}{c|}{100}       &  21.3      &  46.4   &  31.4  &    75.7&  40.8  \\
\multicolumn{3}{c|}{300}      & \textbf{24.3}  & \textbf{47.8}  & \textbf{33.2}  &\textbf{76.3} & \textbf{42.9} \\
\multicolumn{3}{c|}{500}        &    23.6   & 47.0    & 32.7   &75.5    & 42.4   \\
\bottomrule
\end{tabular}
\caption{Impact of minimum instance area threshold during instance refinement on Cityscapes val \cite{cordts2016cityscapes} dataset.}
\label{tab:min-pix-area-ablation}
\vspace{-0.25cm}
\end{table}

We observed that refining the instance segmentation using offsets predicted by center regression marginally improves performance of instance segmentation. However, the refinement is critical in cases where a broken contour can miss the boundary between two instances that can wrongly predicted as a single instances. Similarly, an occlusion by a pole or low width object can mislead connected component labeling to interpret resulting contours as separate instances. 

We observed that choosing an appropriate minimum instance area threshold is critical in determining the performance of our proposed method. Lower instance area allows to remove unwanted instance generated due to artifacts in contour estimation. Such artifacts could be a result of false contours around mirrors of cars, convex hulls, occlusion etc.

\subsubsection{Network Ablation}

We experimented with different network architecture choices as discussed in Section \ref{sec:methods-arch}. We studied the impact of using a shared neck vs separate neck layer to upsample and add features from a common feature pyramid network. We also studied how the depth of ResNet encoder impacts our performance by using  ResNet-50 and ResNet-101 encoders in Table \ref{tab:network-ablation}. We report performance of semantic, instance and panoptic segmentation networks as the change in network architecture impacts learning of different heads. We observed that higher ResNet depth and separate necks yield better performance.

\begin{table}[htb]
\vspace{-0.25cm}
\centering
\begin{tabular}{@{}cc|l|cc|cc|c@{}}
\toprule
\multicolumn{2}{l|}{Neck} & Backbone & mIoU & PQ\textsuperscript{St} & AP & PQ\textsuperscript{Th} & PQ\\ \midrule
\multicolumn{2}{l|}{Shared} & ResNet-50 & 67.5   &  57.4   &  24.3  &  33.2  &  47.8  \\
\multicolumn{2}{l|}{Separate} & ResNet-50  &   \textbf{69.6} & 58.6    & \textbf{25.0}   &  \textbf{34.0}  & 48.3  \\ 
\multicolumn{2}{l|}{Shared} & ResNet-101 &  68.4  &58.5 & 24.7   & 33.4   & 48.1    \\
\multicolumn{2}{l|}{Separate} & ResNet-101  &  68.7 & \textbf{59.3}    &  24.9  & 33.2   & \textbf{48.4}  \\ 
\bottomrule
\end{tabular}
\caption{Performance of semantic, instance and panoptic segmentation using different network architecture choices on Cityscapes val \cite{cordts2016cityscapes} dataset.}
\label{tab:network-ablation}
\vspace{-0.5cm}
\end{table}

\subsection{State of the Art Comparison}
In Table \ref{tab:sota}, we compare our proposed methods against other semantic, instance and panoptic segmentation methods. 

\subsubsection{Comparison with two-stage methods}
As discussed in Section \ref{sec:background-instseg}, two-stage object detection methods \cite{kirillov2019panopticfpn, he2017mask, li2018weakly} dominate the state of the art in instance and panoptic segmentation. However they have incur additional compute costs in generating object detection followed by foreground mask generation. Mask R-CNN \cite{he2017mask} for instance segmentation on a high end GPU like Nvidia Titan X runs at $\sim$5 and $\sim$2 fps on 1024$\times$1024 and 1024$\times$2048 images respectively.

\begin{table}[htb]

\centering
\resizebox{\columnwidth}{!}{
\begin{tabular}{@{}ccc|cc|cc|c@{}}
\toprule
\multicolumn{3}{l|}{Method}  & mIoU & PQ\textsuperscript{St} & AP & PQ\textsuperscript{Th} & PQ\\ \hline \hline
\multicolumn{8}{c}{Two-stage Object detection} \\ 
\hline
\multicolumn{3}{l|}{Mask R-CNN \cite{he2017mask}} & -   &  -   &  31.5  &  -  &  -  \\
\multicolumn{3}{l|}{Weakly Supervised \cite{li2018weakly}}& 71.6   &  52.9   &  24.3  &  39.6  & 47.3  \\
\multicolumn{3}{l|}{Panoptic-FPN \cite{kirillov2019panopticfpn}} & 74.5   &  62.4   &  32.2  &  51.3  &  57.7  \\
\multicolumn{3}{l|}{UPSNet \cite{upsnet}} & 75.2   &  62.7   &  33.3  &  54.6  &  59.3  \\
\multicolumn{3}{l|}{DeepSnake \cite{Peng_2020_CVPR}} & -   &  -   &  37.4  &  -  &  -  \\

\hline \hline
\multicolumn{8}{c}{Instance Clustering} \\
\hline
\multicolumn{3}{l|}{Kendall et al \cite{kendall2018multi} $\dagger$} & 78.5   &  -   &  21.6  &  -  &  -  \\
\multicolumn{3}{l|}{Panoptic-DeepLab \cite{cheng2019panoptic}} & 78.2   &  -   &  32.7  &  -  &  60.3  \\ \hline \hline
\multicolumn{8}{c}{Single-stage Object detection} \\
\hline
\multicolumn{3}{l|}{Poly YOLO
\cite{hurtik2020poly}*} & -   &  -   &  8.7  &  -  &  -  \\\hline \hline
\multicolumn{8}{c}{Others} \\
\hline
\multicolumn{3}{l|}{Deep Contour \cite{deepcontour} $\dagger$} & -   &  -   &  2.3  &  -  &  -  \\
\multicolumn{3}{l|}{Uhrig et al. \cite{uhrig2016pixel}} & -   &  -   &  9.9  &  -  &  -  \\
\multicolumn{3}{l|}{Deep Watershed \cite{bai2017deep}} & -   &  -   &  21.2  &  -  &  -  \\
\multicolumn{3}{l|}{SGN \cite{liu2017sgn}} & -   &  -   &  29.2  &  -  &  -  \\
\hline \hline
\multicolumn{3}{l|}{Ours [ResNet-50]} &  69.6 & 58.6    &  25.0  & 34.0   & 48.3  \\ 
\multicolumn{3}{l|}{Ours [ResNet-101]} &  68.7 & 59.3    &  24.9  & 33.2   & 48.4 \\

\bottomrule
\end{tabular}
}
\caption{Comparison with other state-of-the art methods on Cityscapes \cite{cordts2016cityscapes} dataset (val split). $\dagger$ Performance reported on test split. *\cite{hurtik2020poly} is evaluated on image of size 416$\times$832.}
\label{tab:sota}
\vspace{-0.55cm}
\end{table}

Other two-stage methods UPSNet \cite{upsnet} and DeepSnake \cite{Peng_2020_CVPR} are lighter compared to Mask R-CNN and operate at $\sim$4 fps for instance segmentation task. When semantic segmentation task is executed in parallel with instance segmentation to compute panoptic segmentation the run time speed of these methods will further decline. This makes the two-stage object detection based methods not suitable for real-time applications. Our proposed method with ResNet-50 encoder outputs panoptic segmentation at $\sim$3 fps and $\sim$5 fps on a mid grade Nvidia GTX 1080 GPU ($\sim$8.8 Tflops) on a 1024$\times$2048 image with and without instance refinement function. We expect higher frame rates when our connected component labeling and instance refinement functions are optimized for GPU operation as opposed to its current CPU based implementation.

\subsubsection{Comparison with Instance clustering}
Kendall et al. \cite{kendall2018multi} was one of the early works that used multi-task learning to simultaneously learn semantic and instance segmentation.
Panoptic-DeepLab \cite{cheng2019panoptic} recently proposed an strong baseline for center regression based methods by exploiting the effectiveness of Atrous Spatial Pyramid Pooling (ASPP) modules. We believe that using ASPP module in our network will improve our semantic segmentation performance and eventually lead us to better instance and panoptic segmentation results. However, ASPP modules are computationally very expensive compared to Feature pyramid networks \cite{kirillov2019panopticfpn}. Panoptic-DeepLab with ResNet50 achieves $\sim$5fps on Tesla V-100 SMX2 GPU ($\sim$14 Tflops). 

\subsubsection{Comparison with Single-stage object detection and Others}
Poly YOLO \cite{hurtik2020poly} reported $\sim$22 fps on a 416$\times$832 image with an AP score of 8.7 while Deep Contour \cite{deepcontour} reported $\sim$5fps on a mid grade GTX 1070 with AP score of $2.3^{\dagger}$. Other methods like Deep Watershed \cite{bai2017deep} and SGN \cite{liu2017sgn} ($\sim$0.6 fps) incur a huge computation complexity. Our methods are outperform faster methods like \cite{hurtik2020poly} and \cite{deepcontour} while improving run-time efficiency compared to \cite{bai2017deep, liu2017sgn}.



\section{Conclusion}
\label{sec:conlusion} In this paper, we presented a new approach to panoptic segmentation using instance contours. Our method is one of the first approaches where instance segmentation is a generated as a byproduct in a  semantic segmentation network. We evaluated performance of our semantic, instance and panoptic segmentation results on Cityscapes dataset. We presented several ablation studies that help understand the impact of architecture and training choices that we made. We believe that our proposed methods opens a new direction in research of instance and panoptic segmentation and serves a baseline for contour based methods.
 





\bibliographystyle{unsrt}
\bibliography{egbib}

\end{document}